\documentclass[jimaging,article,accept,pdftex,moreauthors]{Definitions/mdpi} 

\firstpage{1} 
\pubvolume{1}
\issuenum{1}
\articlenumber{0}
\pubyear{2024}
\copyrightyear{2024}
\externaleditor{Academic Editor: Pierre Gouton}
\datereceived{30 March 2024} 
\daterevised{24 April 2024} % Comment out if no revised date
\dateaccepted{4 May 2024} 
\datepublished{ } 
\hreflink{https://doi.org/} % If needed use \linebreak
\Title{A New Dataset and Comparative Study for Aphid Cluster Detection and Segmentation in Sorghum Fields} %Attention: title altered.

\TitleCitation{A New Dataset and Comparative Study for Aphid Cluster Detection and Segmentation in Sorghum Fields}

\Author{{Raiyan} Rahman $^{1}$, Christopher Indris $^{1}$, Goetz Bramesfeld $^{2}$, Tianxiao Zhang $^{3}$, Kaidong Li $^{3}$, Xiangyu Chen $^{3}$, \mbox{Ivan Grijalva} $^{4}$, Brian McCornack $^{4}$, Daniel Flippo $^{5}$, Ajay Sharda $^{5}$ and Guanghui Wang $^{1,}$*}

\AuthorNames{Raiyan Rahman, Christopher Indris, Goetz Bramesfeld, Tianxiao Zhang, Kaidong Li, Xiangyu Chen, \mbox{Ivan Grijalva}, Brian McCornack, Daniel Flippo, Ajay Sharda and Guanghui Wang}

\AuthorCitation{Rahman, R.; Indris, C.; Bramesfeld, G.; Zhang, T.; Li, K.; Chen, X.; Grijalva, I.; McCornack, B.; Flippo, D.; Sharda, A.; et al.}
\address{%
$^{1}$ \quad Department of Computer Science, Toronto Metropolitan University, Toronto, {ON, } %MDPI: 1. We add abbreviation of continent or province, please confirm, followings are same 2. Please add the postal code (or ZIP code in the U.S. and Canada). If the postal code is not available, please provide the P.O. Box. followings are same 
Canada; {raiyan.rahman@torontomu.ca (R.R.); christopher.indris@torontomu.ca (C.I.)} %MDPI: We added the email addresses here according to the submitting system. Please confirm.
\\
$^{2}$ \quad Department of Aerospace Engineering, Toronto Metropolitan University, Toronto, {ON}, Canada;  {bramesfeld@torontomu.ca}\\
$^{3}$ \quad Department of Electrical Engineering and Computer Science, University of Kansas, Lawrence, {KS}, USA; {tianxiao@ku.edu (T.Z.); kaidong.li@ku.edu (K.L.); xychen@ku.edu (X.C.)}\\
$^{4}$ \quad Department of Entomology, Kansas State University, Manhattan, {KS}, USA; {grijalva@ksu.edu (I.G.); mccornac@ksu.edu (B.M.)}\\
$^{5}$ \quad Department of Biological and Agricultural Engineering, Kansas State University, Manhattan, {KS}, USA; {dkflippo@ksu.edu (D.F.); asharda@ksu.edu (A.S.)}
}

\corres{Correspondence: wangcs@torontomu.ca}

\abstract{Aphid infestations are one of the primary causes of extensive damage to wheat and sorghum fields and are one of the most common vectors for plant viruses, resulting in significant agricultural yield losses. To address this problem, farmers often employ the inefficient use of harmful chemical pesticides that have negative health and environmental impacts. As a result, a large amount of pesticide is wasted on areas without significant pest infestation. This brings to attention the urgent need for an intelligent autonomous system that can locate and spray sufficiently large infestations selectively within the complex crop canopies. We have developed a large multi-scale dataset for aphid cluster detection and segmentation, collected from actual sorghum fields and meticulously annotated to include clusters of aphids. Our dataset comprises a total of 54,742 image patches, showcasing a variety of viewpoints, diverse lighting conditions, and multiple scales, highlighting its effectiveness for real-world applications. In this study, we trained and evaluated four real-time semantic segmentation models and three object detection models specifically for aphid cluster segmentation and detection. Considering the balance between accuracy and efficiency, Fast-SCNN delivered the most effective segmentation results, achieving 80.46\% mean precision, 81.21\% mean recall, and 91.66 frames per second (FPS). For object detection, RT-DETR exhibited the best overall performance with a 61.63\% mean average precision (mAP), 92.6\% mean recall, and 72.55 on an NVIDIA V100 GPU. Our experiments further indicate that aphid cluster segmentation is more suitable for assessing aphid infestations than using detection models.}


\begin{document}

%%%%%%%%%%%%%%%%%%%%%%%%%%%%%%%%%%%%%%%%%%
\section{Introduction}

With a growing global population, the increasing demand for food has led to\linebreak widespread innovations in the agricultural industry. Approximately 37\% of the crops grown worldwide are lost to pest damage, and 13\% of this loss is directly attributed to insects~\cite{amiri2019effective}.  The impact is particularly pronounced in staple crops such as rice, wheat, sorghum, and maize, not only posing significant threats to global food security but also impacting various national economies \cite{finegold2019global}, Consequently, there has been a consistent rise in the use of chemical pesticides to combat these pests and maximize yields. The projected expenditure on pesticides reached around USD 107 billion in 2023 and this figure is expected to further increase. The heightened usage not only translates to increased costs for farmers but also underscores the numerous health risks associated with pesticide use. The conventional pest treatments involve applying chemical pesticides uniformly across the entire field using large sprayers in a continuous spray once a specific infestation threshold is reached. However, this treatment approach becomes excessive when pests are only fractionally present. Therefore, a more efficient method is sought after.

\begin{figure}[H]

\begin{adjustwidth}{-\extralength}{0cm}
%\centering %% If there is a figure in wide page, please release command \centering
  \centering
  \includegraphics[width=1\columnwidth]{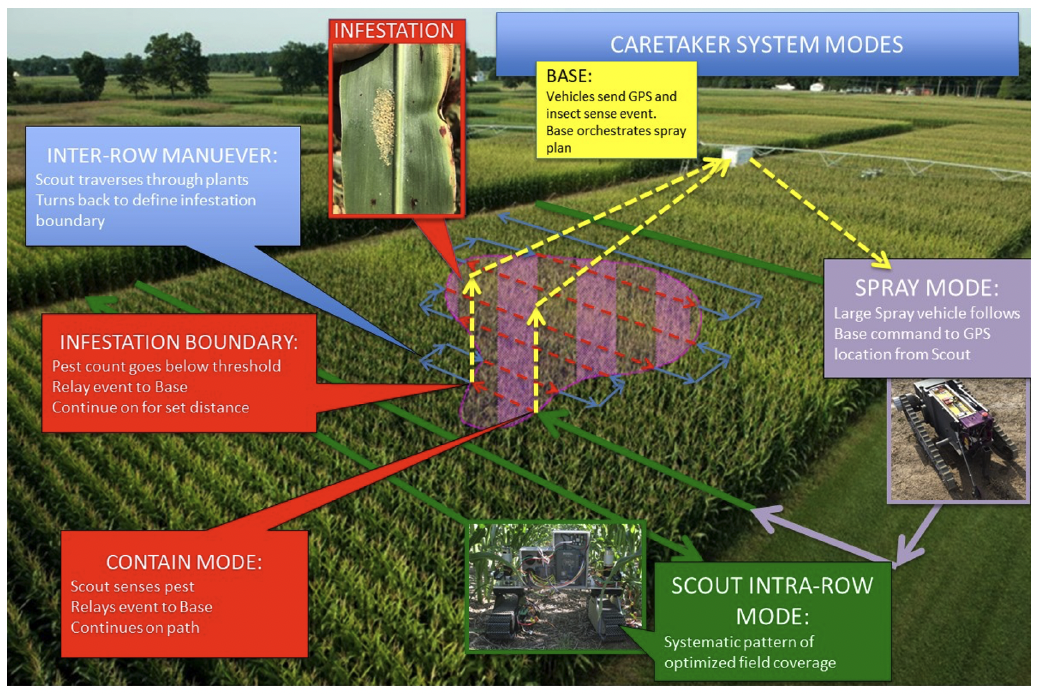}
\end{adjustwidth}
  \caption{An intelligent scouting robot with onboard vision and spray system. The vision system is used to detect and localize aphid affections, and the spray system can apply precise pesticide application to the infected areas.}
  \label{fig:robot}
\end{figure}

In order to provide timely aphid control and reduce the negative impacts of excessive pesticide use through precision spraying, it is desirable to develop an intelligent robot that can scout the field periodically, detect and localize any aphid infections, and apply precise spray once aphids are detected using the on-board spray system, as shown in Figure \ref{fig:robot}. This system will provide a more efficient, effective, and economical solution for pest management. Through data collection and scouting techniques, farmers are able to better plan their crop planting and treatment application to achieve higher yields and save chemical costs by spraying more efficiently. Automated robotic technologies will significantly reduce the labour costs and increase the yield by noticeable margins of up to 5\% \cite{pearce2019}.

The implementation of an automated pest control system holds the potential to significantly enhance farmers' ability to pinpoint areas in need of treatment with precision, thus reducing the overall pesticide usage and fostering sustainable agricultural methods. Despite the extensive research dedicated to automated pest detection, it remains a formidable challenge, primarily due to the small size of individual pests and their adeptness at blending into their natural environment. Previous work has been conducted on the detection and segmentation of aphids \cite{liu2016detection, chen2018automatic, teng2022td, zhang2023new}; however, these works were explored in the context of individual pest detection rather than assessing the overall infestation levels. The datasets are also usually catered to the specific pests and the crops they affect. We explore the feasibility of accurately detecting and segmenting these pests using lightweight models capable of real-time inference speeds.

The primary focus of this paper is on the topic of automated aphid cluster localization through object detection and semantic segmentation tasks. Our multi-scale dataset created using high-resolution images from a sorghum field enables learning to detect aphid clusters at different scales and in robust conditions. We trained popular real-time object detection and semantic segmentation models to evaluate their performance against each other and for the overall application of aphid infestation control. Through a comprehensive analysis, considering the prediction precision, inference speed, and information parameters for automated pest localization, we identify RT-DETR and Fast-SCNN as the top performers for aphid detection and semantic segmentation tasks, respectively. Furthermore, our findings suggest that semantic segmentation proves more advantageous for this application due to the additional spatial information it offers.

The main contributions of this study are summarized below.
\begin{itemize}
    \item We created an extensive multi-scale dataset for aphid cluster detection and semantic segmentation, utilizing real-world images captured in sorghum fields. This dataset encompasses various viewpoints, diverse lighting conditions, and multiple scales, thereby providing enhanced representations of aphid features. By leveraging this dataset, we significantly improve the performance of the learning models in detecting and segmenting aphid clusters.
    \item We conducted a benchmark evaluation to assess the performance of four state-of-the-art real-time semantic segmentation models and three detection models using our generated dataset. The outcomes of this evaluation offer a practical reference point for aphid detection and segmentation in real-world scenarios.
\end{itemize}

Partial results of this paper were previously presented at the CVPR 2023 Workshops~\cite{rahman2023real}. This paper is a significant extension of the workshop paper. In addition to segmentation, the paper incorporates the task of aphid cluster detection with three additional real-time detection models, a comparison with the previous fixed-scale dataset, as well as expanded details regarding the segmentation models. The generated dataset can be downloaded at
 \url{https://doi.org/10.7910/DVN/N3YJXG}.

%-------------------------------------------------------------------------
\section{Related Works}

Convolutional Neural Networks (CNNs) have demonstrated exceptional effectiveness in feature extraction during the past decade. Utilizing CNNs has led to state-of-the-art achievements in various image analysis tasks, including image classification \cite{bur2023interpretable, ma2022semantic}, object detection \cite{li2021colonoscopy, zhang2022dynamic}, and semantic segmentation \cite{he2021sosd, patel22fuzzynet}.
However, the task of pest localization remains challenging for these models, primarily due to the diminutive sizes of insects. Barbedo \cite{barbedo2014using} spearheaded the use of image processing techniques and handcrafted features to automatically detect and count pests on leaves with simple black backgrounds, but more effective features were needed to improve the performance. Using the histogram-oriented gradient algorithm and a support vector machine,  {Ref.} %MDPI: The reference cannot directly be the subject of the sentence. Newly added content, please confirm
 \cite{liu2016detection} devised a way to identify and count aphids with greater accuracy. However, with the advent of CNNs, there was a push to move towards learning features directly from the data.

A modification of the widely used U-Net model was introduced in \cite{chen2018automatic} to segment semantically and count the individual aphid nymphs on leaves. While this method demonstrated high precision and recall, its applicability was limited by the dataset consisting solely of ideal laboratory images featuring simple black backgrounds. Consequently, the model lacked generalizability to the natural settings where these pests are typically found. Hence, it is imperative for models to exhibit robustness to diverse viewpoints and intricate crop canopies in the real-world environments where these systems are deployed. A tiny-sized dense distribution network, TD-Det, was proposed in \cite{teng2022td} and evaluated using the APHID-4K dataset from \cite{DU2022400}. The model consisted of a transformer feature pyramid network and used a multi-resolution training method, highlighting the effectiveness of using multiple resolutions of the images for training. The need for more datasets of this variety was further addressed by the small dataset, LeLePhiD \cite{data6050051}, where they annotated aphid clusters for infestation detection. 

In our preliminary investigation \cite{zhang2023aphid}, we assembled an extensive dataset of aphids in their natural habitat and undertook a comparative analysis of various object detectors using this dataset. However, the dataset was constructed at a fixed scale, and the models implemented were not tailored for real-time applications. Moreover, the bounding boxes provided by the detection models lacked the requisite precision for accurately estimating the infection levels. Given the recent developments regarding real-time models, this paper shifts its focus towards real-time aphid localization within both detection and segmentation contexts.

%-------------------------------------------------------------------------
\section{Dataset}

\begin{figure}[H]
%  \centering
  \includegraphics[width=0.6\columnwidth]{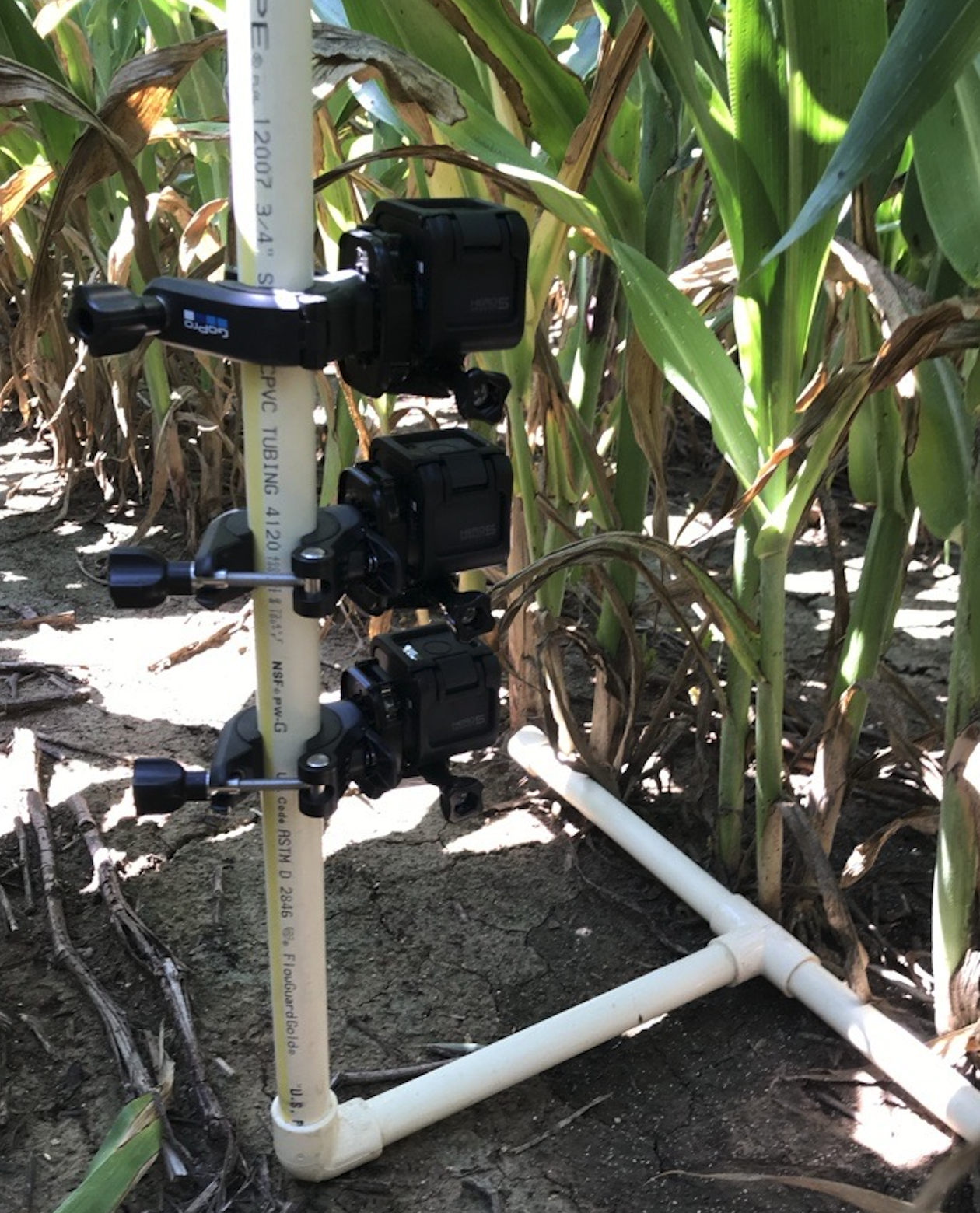}
  \caption{The imaging rig used to capture images. It is equipped with three adjustable GoPro cameras to take images from different heights and viewpoints.}
  \label{fig:camerarig}
\end{figure}

Aphids are recognized for consistently causing significant damage to both grain and sweet sorghums \cite{sorghumaphids}. To ensure the accuracy and representativeness of the data collected for our dataset in capturing the natural habitat of aphids, we took high-resolution images from sorghum fields in both the northern and southern regions of the State of Kansas during the aphid growing season. This was achieved using an imaging rig equipped with three GoPro cameras positioned at different heights, as shown in Figure \ref{fig:camerarig}. This setup facilitates capturing diverse images from various viewpoints, heights, and lighting conditions. This approach enables the models to better comprehend the spatial characteristics of sorghum aphids across different environments, ultimately yielding more robust and generalizable results.

The initial images exhibited sparse aphid concentration, with those devoid of any aphids being filtered out via manual examination by domain-trained research assistants, yielding a dataset comprising 5447 images, each sized at 3647  $\times$  2736 pixels. Given that aphids typically congregate in groups, our approach for infestation assessment focused on identifying clusters rather than individual insects. To this end, during the annotation process, we set a threshold of six or more closely located aphids to define a cluster \cite{zhang2023aphid}. This criterion ensured that only clusters of a significant size, indicative of an economic threat, were detected.

\begin{figure}[H]
 % \centering
  \includegraphics[width=0.8\columnwidth]{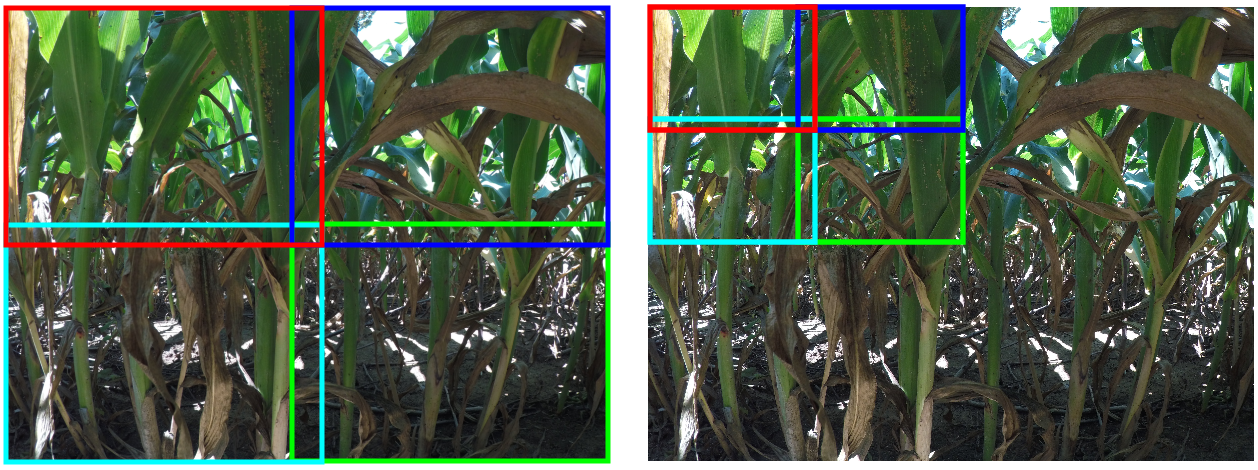}
  \caption{Two examples of the different scales that were used in training. Image (\textbf{left}) shows how the original high-resolution 3647 $\times$ 2736 image was subdivided to create patches at the 0.525W $\times$ 0.525H scale (Scale 3), where W and H refer to the width and height of the original image. At this scale, the original image will yield 4 patches. Image (\textbf{right}) shows how the image was further subdivided to create patches at the 0.263W $\times$ 0.263H scale (Scale 2). This scale will yield 16 patches from the original image. In each case, adjacent patches were taken with an overlap of 10\%.}
  \label{fig:scales}%MDPI: Please provide an explanation for the different colored boxes if you find it necessary. Also, kindly check all figures and tables and ensure that all the special symbols/ formats / colors are explained in the caption. Thank you!
\end{figure}

\begin{figure}[H]
%  \centering
  \includegraphics[width=0.8\columnwidth]{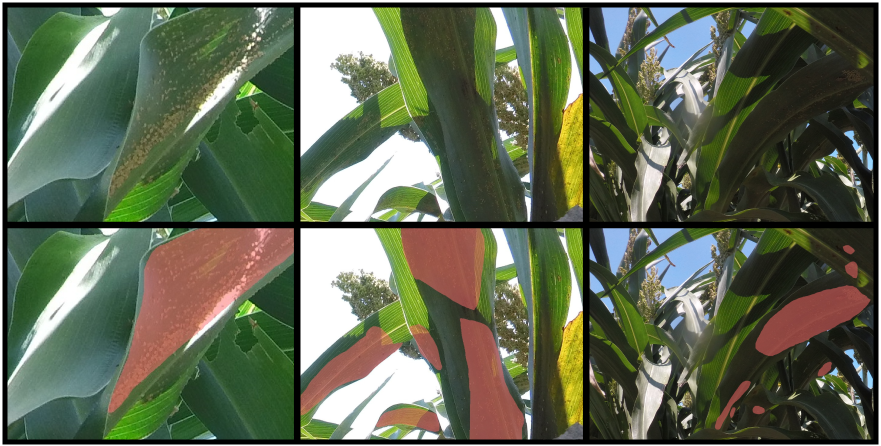}
  \caption{Example images from the dataset alongside their corresponding ground truth labels. The first row shows the appearance of the aphid clusters and the second row has the corresponding ground truth masks overlaid on them. The first, second, and third columns show image patches at Scale 1, Scale 2, and Scale 3, respectively.}
  \label{fig:examples}
\end{figure}

To facilitate a statistically meaningful analysis, we implemented 10-fold cross-validation to mitigate potential biases resulting from random data splitting and the substantial class imbalance against background pixels. The selected images were randomly divided into 10 equal sets. Given the sparsity and small sizes of the aphid clusters in the original high-resolution images, we employed patch generation to enhance cluster presence and ensure robust localization across different scales. This patch generation process, as illustrated in Figure \ref{fig:scales}, involved subdividing the images into patches at three distinct scales: Scale 1 (0.132W  $\times$  0.132H), Scale 2 (0.263W  $\times$  0.263H), and Scale 3 (0.525W  $\times$   0.525H), where W and H represent the width and height of the original image, respectively. Some sample images at different scales are shown in Figure \ref{fig:examples}. To maintain annotation completeness, a 10\% overlap was incorporated between adjacent patches. To prevent any potential data leakage across the sets, the patch generation was conducted after the fold separation of the original images.

\begin{figure}[H]
%\centering
\includegraphics[width=1\textwidth]{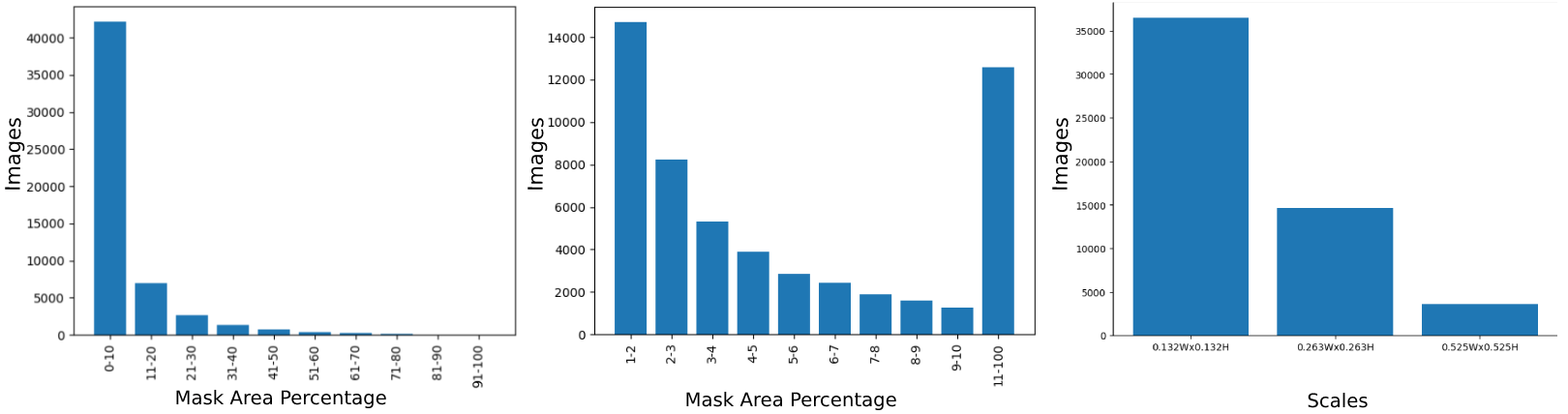}
\caption{ {Histograms} %MDPI: 1. Please use commas to separate thousands for numbers with five or more digits (not for four digits) in the picture. e.g., "10000" should be "10,000" 2. \hl{} %MDPI: Please change the hyphen (-) into endash (--, "U+2013"). e.g., "0-1" should be "0--1".
 showing the mask area percentage across the images and the number of images per scale. The chart on the left shows the percentage of aphid cluster masks using intervals of 10\% from 0\% to 100\%. As most images lie in the interval between 0\% and 10\%, the chart in the center further breaks that interval down for closer analysis. The chart on the right provides the number of images at each scale.}
\label{fig:mask-stats}
\end{figure}

The generated dataset comprised a total of 54,742 multi-scale image patches, with 35,140, 13,311, and 6291 patches captured from the top, middle, and bottom camera heights, respectively. Additionally, there were 36,478, 14,628, and 3636 images obtained from Scale 1, Scale 2, and Scale 3, respectively. Despite the extensive patch generation process, annotations of aphid clusters remained sparse, accounting for only 2.45\% of each image. Initially, annotations were provided in the form of semantic segmentation masks, from which bounding boxes were derived. The statistical distribution of the aphid clusters is shown in Figure \ref{fig:mask-stats}. To create the object detection dataset, overlapping boxes were merged. In contrast, our previous study \cite{zhang2023aphid} exclusively focused on developing the dataset at a single scale. However, as evidenced by our experiments, the incorporation of multi-scale data enables models to learn features at varying resolutions, thereby contributing to improved performance.

In the literature, there are two other aphid image datasets with automated aphids for detection. APHID-4K is a high-resolution dataset containing 4294 images of wheat aphids~\cite{DU2022400}. However, as this dataset is annotated for individual aphid detection and counting, it is less suitable for infestation-level assessment. LeLePhiD is another dataset introduced in \cite{data6050051} for semantically segmenting aphid clusters on lemon leaves. While this dataset is suitable for assessing aphid concentration and finding infestation severity, it is a very small dataset with only 665 images and each image consists of a single lemon leaf taken from 30 to 50 cm away, making it hard for practical applications. Compared to these two datasets, our dataset is significantly larger, with 54,742 images captured at multiple scales and from various viewpoints; it offers a comprehensive and diverse dataset for large-scale pest detection research.

\section{Models}

With the increased demands of practical applications like autonomous vehicles and intelligent systems, efficient models have been developed to quickly process visual information and take immediate action. This highlights the need for models that can work with high-resolution image data at high processing speeds. In this study, we implemented some of the most popular real-time semantic segmentation networks, including Fast-SCNN, BiSeNetV1, BiSeNetV2, and Small HRNet, as well as real-time object detection networks such as Faster R-CNN, RT-DETR, and YOLOv7.

\subsection{Segmentation Models}

Fast-SCNN \cite{DBLP:journals/corr/abs-1902-04502} achieves state-of-the-art segmentation speeds coupled with a \linebreak lightweight design, making it ideal for embedded systems with limited computational memory. Its binary path architecture effectively distributes the computation load and integrates various region-based context information to enhance segmentation accuracy. Bilateral Segmentation Network (BiSeNet) \cite{DBLP:journals/corr/abs-1808-00897, DBLP:journals/corr/abs-2004-02147} is another noteworthy family of real-time segmentation networks employing a similar two-pathway architecture. By fusing high-level semantic information with low-level features, BiSeNet strikes a balance between segmentation speed and accuracy, making it highly suitable for real-time applications. High-Resolution Network (HRNet) \cite{DBLP:journals/corr/abs-1908-07919} serves as a robust backbone for various computer vision tasks. Its simplified version, Small HRNet, achieves real-time speeds while preserving high-resolution representations. This is achieved through a high-resolution stem in the initial stage, followed by the gradual addition of high-to-low-resolution streams in the main body. All these models demonstrate remarkable capability in achieving real-time speeds across different benchmark datasets while maintaining high levels of accuracy. %As such, they are well-suited candidates for our study, addressing the demands of efficient visual processing in diverse applications.

\subsection{Detection Models}

Faster R-CNN is a prominent two-stage real-time object detector building upon the Fast R-CNN model by incorporating a Region Proposal Network (RPN). This RPN shares full-image convolution features with the detection network to enable cost-free region proposals \cite{DBLP:journals/corr/RenHG015}. The YOLO (You Only Look Once) family of models represents state-of-the-art single-stage object detection networks, renowned for their real-time processing speeds and competitive accuracy \cite{DBLP:journals/corr/RedmonDGF15}. YOLOv7, a recent addition to this family, is an anchor-based model capable of achieving higher speed and accuracy compared to other known object detectors. Wang et al. \cite{wang2022yolov7} demonstrated through ablation studies that scaling the network depth and width sequentially, while concatenating the layers, optimizes the YOLOv7 architecture for improved performance. The Real-Time Detection Transformer (RT-DETR), introduced by Lv et al. \cite{lv2023detrs}, stands out as the first real-time end-to-end object detector to achieve state-of-the-art speed and accuracy. It surpasses all YOLO detectors at the same scale by employing a hybrid encoder capable of efficiently processing multi-scale features. RT-DETR also enhances the initialization of object queries through IoU-aware query selection, thereby avoiding delays caused by Non-Maximal Suppression (NMS). Leveraging these three object detectors, we conducted experiments for real-time aphid cluster detection, capitalizing on their respective strengths in speed and accuracy to address the requirements of our study effectively.

%-------------------------------------------------------------------------
\section{Experimental Setup and Evaluation Metrics}

\textbf{ {Pre-Processing:} %MDPI: Please confirm if the bold should be retained. Followings are same
}
The original 3647 $\times$ 2736 images were utilized to generate patches at three different scales. Patches containing less than 1\% aphid cluster coverage were filtered out as they typically represented remnants of clusters from the patch generation process or were deemed insignificant in terms of economic threat. This curation process yielded a final dataset comprising 54,742 multi-scale images. Prior to training, all images were resized to 1024 $\times$ 1024 pixels and normalized based on the mean and standard deviation of the dataset. For the semantic segmentation task, considering the substantial class imbalance between aphid clusters and background classes, class weights were computed using the pixels in the image masks and were applied during training. These weights were calculated as follows:
\begin{equation}
    W_{\textit{Aphid\;Cluster}} = \frac{\textit{Total\;pixels}}{\textit{Aphid\;cluster\;pixels}}
\end{equation}

\begin{equation}
    W_{\textit{Background}} = \frac{\textit{Total\;pixels}}{\textit{Background\;pixels}}
\end{equation}

\textbf{ {Training Setup:}} The models were implemented in Python using PyTorch. For the segmentation models, we utilized the MMSegmentation library \cite{mmseg2020}. Additionally, we employed the MMDetection library for the Faster R-CNN model, while the official implementations were utilized for the other object detection models. The training was conducted using 4 NVIDIA v100 GPUs with a total memory capacity of 64 GB.

\textbf{ {Training Pipeline:}} During training, all segmentation models underwent 160,000 iterations utilizing Stochastic Gradient Descent (SGD) with a learning rate set to 0.001, a momentum of 0.9, a weight decay of 0.005, and a batch size of 2. Conversely, a batch size of 8 images was employed for the object detection models. For training YOLOv7 and Faster R-CNN, SGD was utilized as the optimizer. YOLOv7 was trained with a learning rate of 0.01, a momentum of 0.937, and a weight decay of 0.0005, while Faster R-CNN used a learning rate of 0.02, a momentum of 0.9, and a weight decay of 0.0001. RT-DETR, on the other hand, employed AdamW as the optimizer, trained with a learning rate of 0.0001, a momentum of 0.9, and a weight decay of 0.0001.

\textbf{ {Model Evaluation:}} For the object detection task, we utilized the popular mean Average Precision (mAP) metric across various Intersection over Union (IoU) thresholds, including 0.25, 0.5, and 0.75. In the segmentation task, model evaluation relied on two key metrics: Intersection over Union (IoU) and Dice coefficient. These metrics are defined as below.
\begin{equation}
    IoU = \frac{\textit{Area\;of\;overlap}}{\textit{Area\;of\;union}}
\end{equation}

\begin{equation}
    Dice = \frac{\textit{2} \times \textit{Area of overlap}}{\textit{Total area}}
\end{equation}

The speed of all models is evaluated based on how fast they are able to detect aphid clusters by their frames per second (FPS).

\section{Experimental Results}
Utilizing the generated multi-scale patches, we have achieved a comprehensive representation of aphid clusters from diverse viewpoints, thereby enhancing the robustness of our models across different regions within a sorghum field. By leveraging this dataset, we trained the identified object detection and semantic segmentation networks, thereby establishing a benchmark for aphid cluster localization. This effort contributes significantly to the broader objective of infestation management.

\subsection{Segmentation}

The four real-time semantic segmentation models we used in our experiments included Fast-SCNN, BiSeNetV1, BiSeNetV2, and Small HRNet. The performance of these models is shown in Table \ref{tab:rts-models}. The results show the mean Intersection over Union (mIoU), mean Dice score (mDice), mean precision (mPrecision), mean recall (mRecall), and the speed in frames per second (FPS). From the results, we can see the tradeoff between speed and accuracy, with Small HRNet achieving the highest accuracy at 71.62 mIoU but having the slowest relative speed at 31.57 FPS compared to the other networks. The BiSeNetV1 and BiSeNetV2 models yield lower performance while having quite consistent speeds of 53.70 FPS and 56.19 FPS, respectively. 
In terms of inference speed, Fast-SCNN stands out as the top performer among all the models, achieving above real-time speeds at 91.66 FPS. Remarkably, it maintains a high accuracy level of 71.25 mIoU. Given its superior performance compared to Small HRNet and its exceptional speed, Fast-SCNN emerges as the overall best-performing model for real-time aphid cluster segmentation. Table \ref{tab:rts-models} also shows the overall recommendation in the column ``Rank''.

\begin{table}[H]
  \caption{Real-time segmentation results (sorted via mPrecision). {\textbf{Bold}} indicates top result. ``Rank'' shows the overall recommendation. }
  \label{tab:rts-models}
 % \centering
  \begin{tabularx}{\textwidth}{lclclcl}
    \toprule
    \textbf{Model} & \textbf{mIoU} & \textbf{mDice} & \textbf{mPrecision} & \textbf{mRecall} & \textbf{FPS} & \textbf{Rank}\\
    \midrule
    HRNet-Small & \textbf{71.62 $\pm$ 0.47} & \textbf{81.15 $\pm$ 0.36} & \textbf{80.82 $\pm$ 1.20} & \textbf{81.64 $\pm$ 0.65} & 31.57 & 2\\
    Fast-SCNN & 71.25 $\pm$ 0.59 & 80.87 $\pm$ 0.50 & 80.46 $\pm$ 1.47 & 81.21 $\pm$ 0.67 & \textbf{91.66} &1\\
    BiSeNetV2 & 65.72 $\pm$ 0.53 & 75.58 $\pm$ 0.55 & 77.47 $\pm$ 1.34 & 74.06 $\pm$ 1.05 & 53.70 &3\\
    BiSeNetV1 & 59.94 $\pm$ 0.54 & 69.22 $\pm$ 0.70 & 72.39 $\pm$ 1.55 & 67.12 $\pm$ 1.40 & 56.19 &4\\
    \bottomrule
  \end{tabularx}
\end{table}

\begin{figure}[H]
%\centering
\includegraphics[width=1\textwidth]{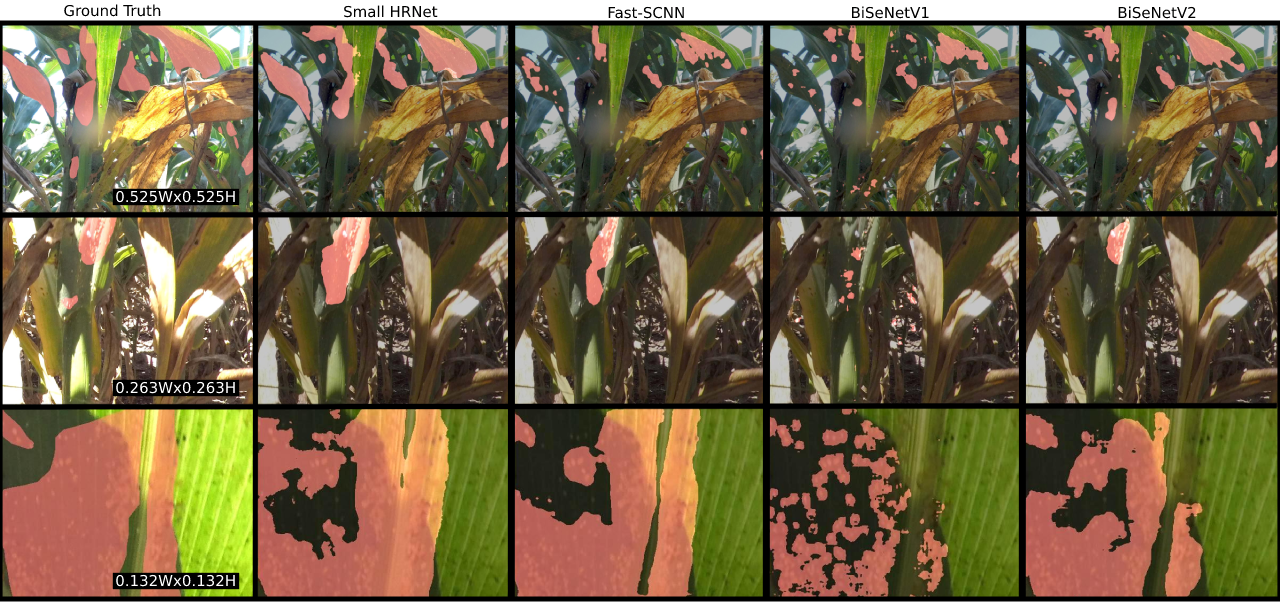}
\caption{The predicted masks from the four real-time semantic segmentation models alongside the ground truth mask annotations. Each row shows us the performance of the models at the three different scales.}
\label{fig:seg-results}
\end{figure}

Figure \ref{fig:seg-results} illustrates the segmentation results across the four models and compares them to the corresponding ground truth annotations. Small HRNet and Fast-SCNN demonstrate the highest accuracy in producing segmentation masks closely aligned with the ground truths. They effectively detect aphid clusters under various visual conditions, including sparsely located clusters, partially occluded clusters, and those in low-light environments. Additionally, Small HRNet exhibits proficiency in handling clusters with more intricate shapes, while Fast-SCNN excels in delineating tighter boundaries, effectively avoiding areas of leaves between clusters devoid of a significant number of aphids. In contrast, the BiSeNet models tend to overlook aphid clusters that blend into the background crop canopy and often misclassify similarly textured parts of the plant as clusters. Notably, BiSeNetV1 may also erroneously detect smaller individual clusters within larger ones rather than recognizing them all as a single cluster.

\subsection{Detection}

The experiments incorporated three real-time object detection models: Faster R-CNN, YOLOv7, and RT-DETR. Table \ref{tab:rtd-models-detailed} presents the performance metrics of these models corresponding to various IoU thresholds of 0.25, 0.5, and 0.75. It is evident that the IoU threshold plays a significant role in determining the difficulty for models to generate True Positive predictions. Higher IoU thresholds impose stricter localization requirements, leading to lower average precision and recall. This underscores the tradeoff between the stringency of the localization criteria and the overall performance of the model. Table \ref{tab:rtd-models} shows the mean average precision and recall averaged across the three IoU thresholds, as well as the corresponding frames per second (FPS) rates, where the column ``Rank'' indicates the overall recommendation for practical deployment.

Among these detection models, RT-DETR achieved the highest accuracy, attaining a 61.63 mAP, while also managing the second fastest inference speed with an above real-time speed of 72.55 FPS. Faster R-CNN emerges as the second-best performer with an accuracy of 57.83 mAP. Even though this model has the slowest relative speed, it is still able to maintain above real-time speeds of 48.03 FPS. YOLOv7 achieved the lowest relative accuracy of 57.33 mAP, which is quite similar to Faster R-CNN, but its lightweight architecture enables it to be more efficient, with the highest frame rate at 113.64 FPS. Although YOLOv7 demonstrates the fastest frame rate, its lower accuracy highlights the tradeoff between speed and accuracy. Consequently, RT-DETR emerges as the optimal choice overall, offering the best balance between accuracy and above real-time inference speeds.

\begin{figure}[H]
%\centering
\includegraphics[width=.99\textwidth]{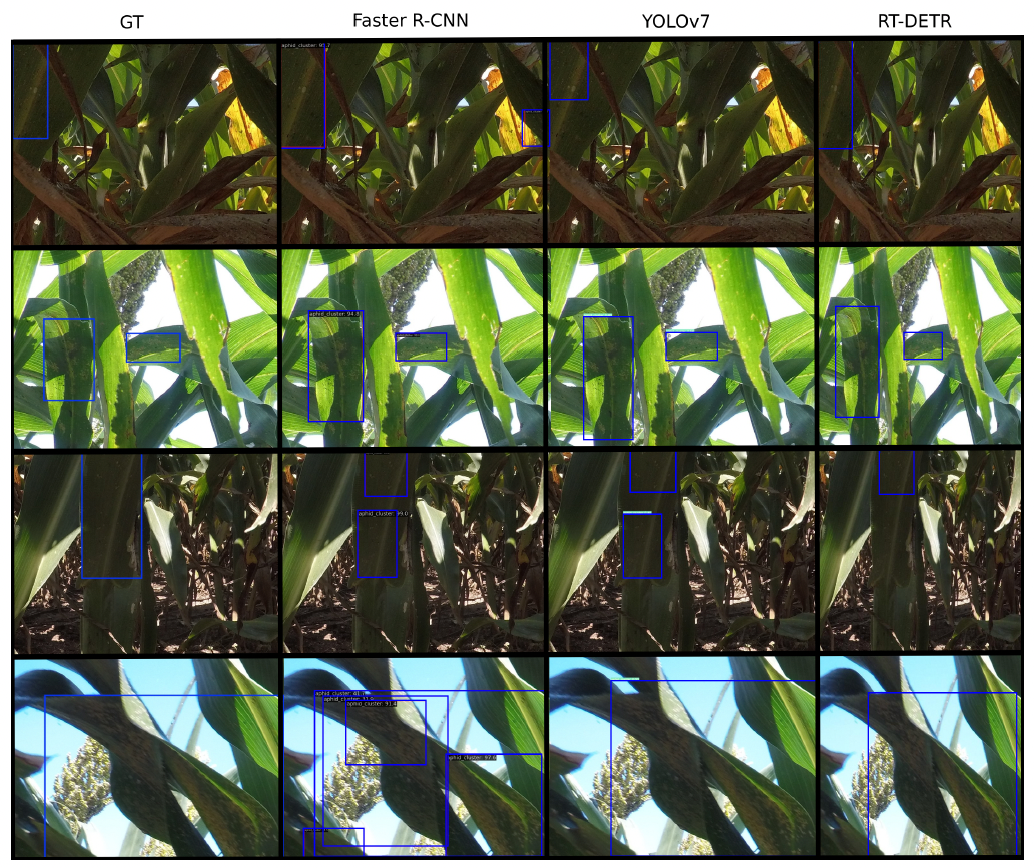}
\caption{ {The} %MDPI: The image content is blocked, please confirm whether it will affect reading.
 predicted bounding boxes from the three object detection models as well as the ground truth bounding boxes. Each row highlights the performance of the models at different lighting conditions and viewpoints.}
\label{fig:det-results}
\end{figure}

The visual detection results from each of the three models, alongside their corresponding ground truth bounding boxes, are depicted in Figure \ref{fig:det-results}.  From these results, we can see that all the models generated bounding boxes that closely aligned with their ground truths. Faster R-CNN, however, tended to produce additional bounding boxes, often detecting smaller clusters in the background that pose a negligible economic threat, and occasionally misidentifying the background components of the crop as clusters. Moreover, due to the absence of Non-Maximal Suppression, Faster R-CNN resulted in overlapping bounding boxes. Despite encountering challenges in detecting the sparsely populated clusters in the dark and shadow-covered bases of the crops, RT-DETR demonstrated robustness across various conditions, including different viewpoints, lighting conditions, leaf sizes, as well as images with blur and shadow.

\begin{table}[H]
  \caption{Object detection results detailed at different IoU thresholds of 0.25, 0.50, and 0.75. For YOLOv7, 0.25, 0.5, and 0.75 were also set as the Non-Maximal Suppression thresholds for the corresponding IoU thresholds.}
  \label{tab:rtd-models-detailed}
%  \centering
  \begin{tabularx}{\textwidth}{lCCC}
    \toprule
    \textbf{Model} & \textbf{Threshold} & \textbf{AP} & \textbf{Recall}\\
    \midrule
    \multirow{3}{*}{Faster R-CNN} & 0.25 & 75.8 & 94.0\\
    & 0.50 & 63.7 & 86.4\\
    & 0.75 & 34.0 & 54.8\\
    \midrule
    \multirow{3}{*}{YOLOv7} & 0.25 & 74.9 & 71.1\\
    & 0.50 & 61.0 & 58.2\\
    & 0.75 & 36.1 & 40.0\\
    \midrule
    \multirow{3}{*}{RT-DETR} & 0.25 & 76.9 & 99.7\\
    & 0.50 & 66.0 & 99.0\\
    & 0.75 & 42.0 & 79.1\\
  %  \midrule
    \midrule
    \multirow{3}{*}{VFNet} & 0.25 & 51.2 & 89.0\\
    & 0.50 & 38.4 & 79.0\\
    & 0.75 & 16.2 & 37.0\\
    \midrule
    \multirow{3}{*}{GFLV2} & 0.25 & 51.3 & 88.2\\
    & 0.50 & 38.2 & 76.6\\
    & 0.75 & 15.5 & 34.5\\
    \midrule
    \multirow{3}{*}{PAA} & 0.25 & 49.2 & 91.6\\
    & 0.50 & 37.9 & 82.4\\
    & 0.75 & 16.5 & 38.8\\
    \midrule
    \multirow{3}{*}{ATSS} & 0.25 & 51.5 & 89.3\\
    & 0.50 & 38.5 & 78.5\\
    & 0.75 & 15.7 & 35.7\\
 %   \midrule
    \bottomrule
  \end{tabularx}
\end{table}

\begin{table}[H]
  \caption{Real-time object detection results (sorted via mAP). The mAP and mRecall are averaged over IoU thresholds at 0.25, 0.50, and 0.75. {\textbf{Bold}} indicates top result. ``Rank'' shows the overall recommendation. ``$\times$'' means not recommended.}
  \label{tab:rtd-models}
 % \centering
  \begin{tabularx}{\textwidth}{lClCl}
    \toprule
    \textbf{Model} & \textbf{mAP} & \textbf{mRecall} & \textbf{FPS}  & \textbf{Rank}\\
    \midrule
    RT-DETR & \textbf{61.63} & \textbf{92.60} & 72.55 &1\\
    Faster R-CNN & 57.83 & 78.40 & 48.03 &3\\
    YOLOv7 & 57.33 & 56.43 & \textbf{113.64} &2\\
    \midrule
    VFNet & 35.27 & 68.33 & 22.00 &  $\times$ \\
    GFLV2 & 35.00 & 66.43 & 21.98 &  $\times$ \\
    PAA & 34.53 & 70.93 & 10.43 &  $\times$ \\
    ATSS & 35.23 & 67.83 & 19.04 &  $\times$ \\
    \bottomrule
  \end{tabularx}
\end{table}

In Table \ref{tab:rtd-models-detailed} and Table \ref{tab:rtd-models}, we also compared the detection performance with the following four detection models reported in \cite{zhang2023aphid}: ATSS (Adaptive Training Sample Selection) \cite{zhang2020bridging}, GFLV2 (Generalized Focal Loss V2) \cite{li2021generalized}, PAA (Probabilistic Anchor Assignment) \cite{kim2020probabilistic}, and VFNet (VarifocalNet) \cite{zhang2021varifocalnet}. Although these models were trained and evaluated from the same original image dataset, their performance does not align with the results obtained in this study. The primary reason for this variance is that these models were trained using the fixed-scale dataset created in \cite{zhang2023aphid}. This comparison further underscores the significance of creating a multi-scale dataset, emphasizing its crucial role in achieving superior performance in aphid cluster detection and segmentation tasks.

%-------------------------------------------------------------------------
\section{Discussion}

\subsection{Results Analysis}

Despite the advancements in modern computer vision techniques, localizing small insects remains a very challenging problem in automated detection and segmentation within the agricultural field. This difficulty primarily arises from the minuscule sizes of these pests and their adeptness at blending into complex natural crop environments. To tackle these obstacles, we implemented a three-camera setup with varying heights and employed multi-scale patch generation to create object detection and semantic segmentation datasets. As a result, the resulting images encompassed a wide range of scales, viewpoints, and heights, surpassing the diversity of the existing pest image datasets. Moreover, recognizing that aphid infestations are typically clustered, with the size of the cluster indicating the level of economic threat, our annotations occurred at the cluster level rather than focusing on individual insects. Additionally, the utilization of class weights helped to address the significant class imbalance, ensuring that the segmentation models could effectively learn from the data.

Among the models evaluated, RT-DETR emerged as the top performer for object detection, while Fast-SCNN excelled in semantic segmentation. This success can be attributed to their combination of high speed and accuracy, making them well-suited for practical agricultural applications. With its newly designed hybrid encoder able to efficiently process multi-scale features, RT-DETR is able to avoid the delays caused by NMS and reduce the computation time of the network. Using IoU-aware query selection also allows the model to focus on the parts of the image that contain the most relevant objects, enhancing the accuracy. From the results in Figure \ref{fig:det-results}, we can see that RT-DETR performs quite well relative to the other models and is robust to changes in lighting conditions and viewpoints. Although YOLOv7 manages to achieve a higher frame rate, the combination of the highest accuracy at a speed above real time ensures that RT-DETR is sufficient for quality real-time aphid cluster detection with higher recall. From Table \ref{tab:rtd-models-detailed}, we also notice the tradeoff between the localization strictness and the model's performance. As the agricultural needs are vast and diverse, this enables the threshold to be customized to different requirements.

The highly recommended model, Fast-SCNN, has a binary path architecture, which allows it to efficiently downsample the images and concatenate the shared low-level computed features to achieve its excellent speeds. However, it is this simplicity that keeps it from outperforming the most accurate model, Small HRNet, which sacrifices speed for its better accuracy. We see this further demonstrated in Figure \ref{fig:seg-results} where Small HRNet is able to segment more complicated boundaries to better fit the aphid clusters. Fast-SCNN is still, however, able to quite accurately segment the clusters and account for different lighting conditions and complicated crop backgrounds.

\subsection{Recommendation for Aphid Infestation Control}

Given that aphids typically appear sparsely but form dense clusters, the economic impact of these pests correlates directly with the size of these clusters. For an autonomous pest control system to effectively manage infestations in real time, it must quickly localize these clusters and assess the severity of the infestation based on their sizes. Our comparative results of object detection and semantic segmentation approaches demonstrate that the additional spatial information provided by semantic segmentation, through the sizes of the predicted masks, allows for an accurate determination of whether an infestation has reached a critical threshold requiring intervention.

\begin{figure}[H]
%\centering
\includegraphics[width=0.7\textwidth]{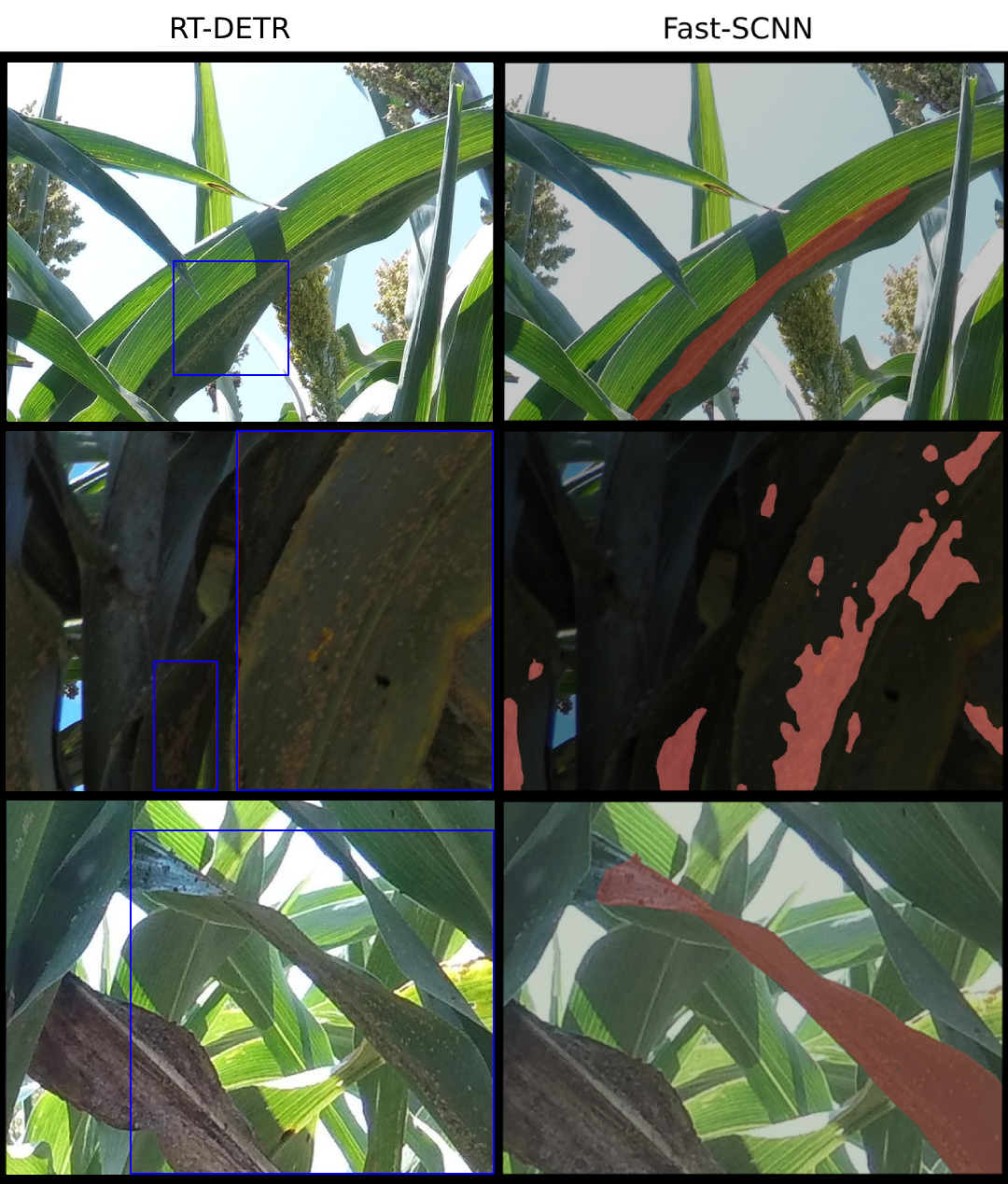}
\caption{Prediction results for the same test images in object detection and semantic
segmentation contexts from RT-DETR and Fast-SCNN, respectively. These examples show how semantic segmentation is more beneficial due to the provided spatial information of the aphid clusters.}
\label{fig:detseg-comparison}
\end{figure}

From Figure \ref{fig:detseg-comparison}, we observe the predictions achieved by RT-DETR and Fast-SCNN for object detection and semantic segmentation, respectively, on identical test images. While both methodologies are adept at identifying small, isolated, and circular clusters of aphids, challenges arise with atypical cluster formations. For example, in cases where clusters form elongated lines along leaves or assume elongated shapes, as illustrated in the bottom row of Figure \ref{fig:detseg-comparison}, bounding boxes may cover extensive areas of the image, yet the actual clusters occupy only sparse sections within these boxes. Conversely, the predicted masks from semantic segmentation provide a more accurate representation of the area directly occupied by the aphid clusters.

Therefore, we recommend semantic segmentation for this task as its pixel-level precision in identifying aphid clusters proves to be more suitable. While RT-DETR provides high accuracy at speeds above real time, Fast-SCNN’s capability to accurately segment clusters with precise boundaries offers a more accurate assessment of the infestation level present in the image. Nonetheless, we recognize that agricultural settings vary widely, and both methodologies have demonstrated effectiveness in localizing aphid clusters in their natural habitats. In scenarios where speed is a priority, or the sheer number of detected clusters is sufficient for deciding on treatment actions, the detection approach may be preferred, still yielding competitive results. In practical applications, we might establish a threshold such that, if the total area covered by segmentation masks or detection bounding boxes in an image exceeds a predetermined threshold, the onboard spray system would be activated to apply pesticides to the identified areas.

%-------------------------------------------------------------------------
\section{Conclusion}

Aphids cause some of the most consistent damage to wheat, rice, and sorghum crops worldwide. As a primary cause of insect pest damage, recent advances in automated drone technology have allowed farmers to target them more effectively using selective treatment. However, for computer vision models to be able to detect and segment aphid clusters with a sufficient level of performance, they require a large amount of diverse data. Using manually collected and labelled high-resolution images, our large multi-scale dataset of aphid clusters addresses the need to adequately assess the infestation levels of aphids in sorghum crops. The different scales allow for better generalization capabilities and robustness to different ranges of zoom and viewpoints. In our experiments, we trained popular real-time object detection and semantic segmentation models to localize aphid clusters regarding real-time speeds or better. Our results show that both RT-DETR and Fast-SCNN are ideal choices for object detection and semantic segmentation, respectively, for providing high-quality predictions at above real-time speeds. In terms of the overall better-suited task, semantic segmentation is recommended due to the extra spatial information provided through the masks. These findings highlight the promise of this problem domain, and further studies with deployed models in the field may enable real-world data on the effects of these models on the yield to be studied. Thus, this study will hopefully motivate further research that drives us towards a more sustainable and efficient agricultural system.

%%%%%%%%%%%%%%%%%%%%%%%%%%%%%%%%%%%%%%%%%%
%\section{Patents}

%This section is not mandatory, but may be added if there are patents resulting from the work %reported in this manuscript.

%%%%%%%%%%%%%%%%%%%%%%%%%%%%%%%%%%%%%%%%%%
\vspace{6pt} 

%%%%%%%%%%%%%%%%%%%%%%%%%%%%%%%%%%%%%%%%%%
%% optional
%\supplementary{The following supporting information can be downloaded at:  \linksupplementary{s1}, Figure S1: title; Table S1: title; Video S1: title.}

% Only for journal Methods and Protocols:
% If you wish to submit a video article, please do so with any other supplementary material.
% \supplementary{The following supporting information can be downloaded at: \linksupplementary{s1}, Figure S1: title; Table S1: title; Video S1: title. A supporting video article is available at doi: link.}

% Only for journal Hardware:
% If you wish to submit a video article, please do so with any other supplementary material.
% \supplementary{The following supporting information can be downloaded at: \linksupplementary{s1}, Figure S1: title; Table S1: title; Video S1: title.\vspace{6pt}\\
%\begin{tabularx}{\textwidth}{lll}
%\toprule
%\textbf{Name} & \textbf{Type} & \textbf{Description} \\
%\midrule
%S1 & Python script (.py) & Script of python source code used in XX \\
%S2 & Text (.txt) & Script of modelling code used to make Figure X \\
%S3 & Text (.txt) & Raw data from experiment X \\
%S4 & Video (.mp4) & Video demonstrating the hardware in use \\
%... & ... & ... \\
%\bottomrule
%\end{tabularx}
%}

%%%%%%%%%%%%%%%%%%%%%%%%%%%%%%%%%%%%%%%%%%
\authorcontributions{Conceptualization: B.M., D.F., A.S., and G.W.; methodology and experiment analysis: R.R., C.I., G.B., T.Z., and G.W.; data curation: K.L.,  {X.C.,} 
 and I.G.; writing and editing: R.R., C.I., T.Z., and G.W. 
All authors have read and agreed to the published version of the manuscript.}

\funding{This research was partly funded by the Natural Sciences and Engineering Research Council of Canada (NSERC) under grant no. RGPIN2021-04244 and the United States Department of Agriculture (USDA) under grant no. 2019-67021-28996.}

\institutionalreview{Not applicable.}

\informedconsent{{Not applicable. } %Any research article describing a study involving humans should contain this statement. Please add “Informed consent was obtained from all subjects involved in the study.” OR “Patient consent was waived due to REASON (please provide a detailed justification).” OR “Not applicable” for studies not involving humans. You might also choose to exclude this statement if the study did not involve humans.
}
%Written informed consent for publication must be obtained from participating patients who can be identified (including by the patients themselves). Please state ``Written informed consent has been obtained from the patient(s) to publish this paper'' if applicable.}

\dataavailability{The dataset generated in this study can be downloaded at
 \url{https://doi.org/10.7910/DVN/N3YJXG}.} 

% Only for journal Nursing Reports
%\publicinvolvement{Please describe how the public (patients, consumers, carers) were involved in the research. Consider reporting against the GRIPP2 (Guidance for Reporting Involvement of Patients and the Public) checklist. If the public were not involved in any aspect of the research add: ``No public involvement in any aspect of this research''.}

% Only for journal Nursing Reports
%\guidelinesstandards{Please add a statement indicating which reporting guideline was used when drafting the report. For example, ``This manuscript was drafted against the XXX (the full name of reporting guidelines and citation) for XXX (type of research) research''. A complete list of reporting guidelines can be accessed via the equator network: \url{https://www.equator-network.org/}.}

% Only for journal Nursing Reports
%\useofartificialintelligence{Please describe in detail any and all uses of artificial intelligence (AI) or AI-assisted tools used in the preparation of the manuscript. This may include, but is not limited to, language translation, language editing and grammar, or generating text. Alternatively, please state that “AI or AI-assisted tools were not used in drafting any aspect of this manuscript”.}

%\acknowledgments{The authors would like to thank many researchers at the University of Kansas and Kansas State University for their help in the collection and annotation of the dataset, especially to Xiangyu Chen, Krushi Patel, and Mateus Raitz.}

\conflictsofinterest{The authors declare no conflicts of interest.} 

\begin{adjustwidth}{-\extralength}{0cm}
%\printendnotes[custom] % Un-comment to print a list of endnotes

\reftitle{References}

\PublishersNote{}
\end{adjustwidth}
\end{document}